\documentclass[10pt,twocolumn,letterpaper]{article}

\usepackage{cvpr}
\usepackage{times}
\usepackage{epsfig}
\usepackage{graphicx}
\usepackage{amsmath}
\usepackage{amssymb}
\usepackage{subcaption}
\usepackage{color}
\usepackage{multirow}
\usepackage{array}
\definecolor{graycolor}{rgb}{0.55,0.55,0.55} 

\usepackage{epstopdf}

\usepackage[pagebackref=true,breaklinks=true,letterpaper=true,colorlinks,bookmarks=false]{hyperref}

\cvprfinalcopy 

\def\cvprPaperID{2016} 

\ifcvprfinal\fi
\begin{document}

\title{Global Second-order Pooling Convolutional Networks}

\author{Zilin Gao$^{\dagger}$, Jiangtao Xie$^{\dagger}$, Qilong Wang$^{\ddagger}$, Peihua Li$^{\dagger}$\\
$^{\dagger}$Dalian University of Technology, $^{\ddagger}$Tianjin University\\
{\tt\small peihuali@dlut.edu.cn}
}

\maketitle

\begin{abstract}
Deep Convolutional Networks (ConvNets) are fundamental to, besides  large-scale visual recognition, a lot of  vision tasks. As the primary goal of the ConvNets is  to characterize complex  boundaries of thousands of classes in a high-dimensional space, it is critical to learn higher-order representations for enhancing non-linear modeling capability.  Recently, Global Second-order Pooling (GSoP), plugged at the end of networks, has attracted increasing attentions, achieving much better performance than classical, first-order networks  in a variety of vision tasks. However, how to effectively introduce higher-order representation in earlier layers for improving non-linear capability of ConvNets  is still an open problem. In this paper, we propose a novel network model introducing GSoP across from lower to higher layers for exploiting holistic image information throughout a network. Given an input 3D tensor outputted by some previous convolutional  layer, we perform  GSoP to obtain a covariance matrix which, after nonlinear transformation, is  used for tensor scaling along channel dimension. Similarly, we can perform  GSoP along spatial dimension for tensor scaling as well. In this way, we can make full use of the second-order statistics  of the holistic image throughout a network.  The proposed networks are thoroughly  evaluated on large-scale ImageNet-1K, and experiments have shown that they outperformed non-trivially  the counterparts while achieving state-of-the-art results.
\end{abstract}

\section{Introduction}

Deep Convolutional  Networks (ConvNets)  are fundamental to computer vision field, since they are  not only paramount  for high accuracy of large-scale object recognition, but also play central roles, through means of pretrained models, in   advancing  substantially many other computer vision tasks,  e.g., object detection~\cite{Redmon_2017_CVPR}, semantic segmentation~\cite{Long_2015_CVPR} and video classification~\cite{Wang_2018_CVPR_nonlocal}.  
Given color images as inputs, the ConvNets can learn progressively the low-level, mid-level and high-level features~\cite{Zeiler2014}, finally producing  global image representations connected to  soft-max layer for classification. To better characterize complex boundaries of thousands of classes in a very high-dimensional space, one possible solution is to learn higher-order representations for enhancing nonlinear modeling capability of ConvNets. 

Recently, modeling of higher-order statistics for more discriminative image representations  has attracted great interests in deep ConvNets. The global second-order pooling (GSoP), producing covariance matrices as image representations, has achieved state-of-the-art results in a variety of vision tasks~\cite{Li_2018_CVPR,Cui_2017_CVPR,Wang_2018_CVPR,WangYunbo_2017_CVPR} such as  object recognition, fine-grained visual categorization, object detection and video classification.  The pioneering  works, i.e., DeepO$_{2}$P~\cite{Ionescu_2015_ICCV} and  bilinear CNN (B-CNN)~\cite{lin2015bilinear}, performed  global second-order pooling, rather than the commonly used global average (i.e., first-order) pooling (GAvP)~\cite{Lin_ICLR_2014_GFoP},  after the last convolutional layers in an end-to-end manner. However, most of the variants of GSoP~\cite{Gao_2016_CVPR,Cai_2017_ICCV} only focused on small-scale scenarios. In large-scale visual recognition,  MPN-COV~\cite{Li_2017_ICCV,Li_2018_CVPR} has shown matrix power normalized GSoP can significantly  outperform global average pooling.

Though GSoP plugged at the end of network  has proven successful,  how to effectively introduce higher-order representation in earlier layers for improving non-linear capability of ConvNets is still an open problem. Several works~\cite{LiYanghao_2017_ICCV,Wang_2017_ICCV,Zoumpourlis_2017_ICCV} have made attempts  to enhance non-linear
modeling capability using quadratic transformation to model feature interactions, instead of only using linear transformation of convolutions.  However, performance gains of these
methods are limited in large-scale visual recognition.   Motivated by Squeeze-and-Excitation (SE) networks~\cite{Hu_2018_CVPR_SE}, we introduce GSoP across from lower to higher layers of deep ConvNets,  aiming to learn more discriminative  representations  by exploiting the second-order statistics of holistic image throughout a deep ConvNet.

At the heart of our global second-order networks is the GSoP block,  which can be conveniently plugged into any location of a deep ConvNet.  Given a 3D  tensor outputted by some previous convolutional  layer, we first perform  GSoP to model pairwise  channel correlations of the holistic tensor. We then accomplish embedding of the resulting covariance matrix by convolutions and non-linear activations, which is finally used for scaling the 3D tensor along  channel dimension. The diagram of our GSoP convolutional network (GSoP-Net) is presented in Figure~\ref{subfigure:overview} and the proposed second-order block is illustrated in Figure~\ref{subfigure:GSoP-block}. The primary differences of the proposed GSoP-Net from existing networks are compared in Table~\ref{tab:overall}, which will be detailed in next section. Our main contributions are threefold.  (1) Distinct from the existing methods which can only exploit second-order statistics at network end, we are among the first who  introduce  this modeling into intermediate  layers for making use of holistic image information in earlier stages of deep ConvNets.   By modeling the correlations of  the holistic tensor, the proposed blocks can capture  long-range statistical dependency~\cite{Wang_2018_CVPR_nonlocal}, making full use of the contextual information in the image. (2) We design a simple yet effective GSoP block, which is highly modular with  low memory and computational complexity.  The GSoP block, which is able to capture global second-order statistics along channel dimension or position dimension,  can be conveniently  plugged into existing network architectures, further improving their performance with small overhead. (3)  On ImageNet benchmark, we perform a thorough ablation study of the proposed  networks, analyzing the characteristics and behaviors of the proposed GSoP block. Extensive comparison with the counterparts has shown the competitiveness of our networks.
 
\section{Related Works}\label{section:related-work}

\paragraph{GAvP (1$^{\mathrm{st}}$--order) In-between Network. }  Global average pooling plugged at the end of network~\cite{Lin_ICLR_2014_GFoP}, which summarizes the first-order statistics (i.e., mean vector) as image representations,  has been widely used in most deep ConvNets such as ResNet~\cite{He_2016_CVPR}, Inception~\cite{Szegedy_2015_CVPR} and DenseNet~\cite{Huang_2017_CVPR}. For the first time, SE-Net~\cite{Hu_2018_CVPR_SE} introduced GAvP in-between network for making use of holistic image context at earlier stages, reporting significant  improvement over its network-end counterparts. The SE-Net  consists of two modules: a squeeze module accomplishing  global average pooling followed by convolution and non-linear activations for  capturing  channel dependency, and an excitation module scaling channel for data recalibration.  Besides GAvP along channel dimension, CBAM~\cite{Woo_2018_ECCV} extends  the idea of SE-Net, combining  GAvP along channel dimension as well as spatial dimension for accomplishing  self-attention.   Compared to SE-Net and CBAM which uses only first-order statistics (mean) of the holistic  image, our GSoP-Net exploits second-order statistics (correlations), having stronger modeling capability.

\paragraph{GSoP (2$^{\mathrm{nd}}$--order) at Network Net.}

The global second-order pooling, plugged at network end and trainable in an end-to-end manner, has received great interests, achieving significant  performance improvement~\cite{Cui_2017_CVPR,Li_2017_ICCV,Li_2018_CVPR}. Several researchers~\cite{Gao_2016_CVPR,Cui_2017_CVPR,Cai_2017_ICCV} have shown close connections between higher-order pooling with kernel machines, based on which they proposed explicit  mapping functions as kernel approximation  for compactness of covariance representations. Wang et al.~\cite{Wang_2017_CVPR} proposed a global Gaussian distribution embedding network (G$^2$DeNet), where one multivariate Gaussian, identified as a symmetric positive definite matrix of covariance matrix and mean vector~\cite{LE2MG}, is plugged at network end. MoNet~\cite{Xiao_2018_CVPR} proposed a sub-matrix square-root layer, making G$^2$DeNet to have compact representation. In~\cite{Dai_2017_CVPR}, the first-order information are combined with the second-order one which achieves consistent  improvements over the standard bilinear  networks on texture recognition.  In all the aforementioned works, second-order modeling are only exploited at the end of deep networks. 

\begin{table}[t]
	\centering
	\setlength{\tabcolsep}{2pt}
	\renewcommand{\baselinestretch}{1.1}
	\small
	\renewcommand\arraystretch{1.2}
	\begin{tabular}{c|cc|cc}
		\hline	
		& \multicolumn{2}{c|}{in-between network} & \multicolumn{2}{c}{end of network} \\
		\cline{2-5} 
		& global pool & means & global pool & means \\
		\hline
		\parbox{0.8in}{\vspace{1mm}
			AlexNet~\cite{Krizhevsky2012ImageNet}\\ VGG~\cite{Simonyan15}\vspace{1mm}} & $\times$  & N$/$A  & $\times$ & N$/$A \\
		\hline
		\parbox{0.8in}{\vspace{1mm}ResNet~\cite{He_2016_CVPR} \\Inception~\cite{Szegedy_2015_CVPR} DenseNet~\cite{Huang_2017_CVPR}\vspace{1mm}} & $\times$ & N$/$A & $\surd$ & 1$^{\mathrm{st}}$--order \\
		\hline
		\parbox{0.85in}{\vspace{1mm}SE-Net~\cite{Hu_2018_CVPR_SE}\\CBAM~\cite{Woo_2018_ECCV}\vspace{1mm}} & $\surd$ & ${1^\mathrm{st}}$--order & $\surd$ & ${1^\mathrm{st}}$--order\\ 
		\hline
		\parbox{0.85in}{\vspace{1mm}DeepO$_{2}$P~\cite{Ionescu_2015_ICCV}\;$\;$ B-CNN~\cite{lin2015bilinear} \\MPN-COV~\cite{Li_2017_ICCV} G$^{2}$DeNet~\cite{Wang_2017_CVPR}\vspace{1mm}} & $\times$ & N$/$A & $\surd$ & 2$^{\mathrm{nd}}$--order \\
		\hline
		GSoP-Net (ours) & $\surd$ & 2$^{\mathrm{nd}}$--order & $\surd$  & 2$^{\mathrm{nd}}$--order \\
		\hline	
	\end{tabular}%
	\renewcommand{\baselinestretch}{1.0}    \vspace{0pt}
	\caption{Summary of  ConvNet models in terms of global statistical  pooling. Different from existing networks, we introduce global second-order pooling into intermediate layers of deep ConvNets. So we can make full use of second-order statistics to effectively capture holistic image information throughout a network.}
	\label{tab:overall}%
\end{table}

  \begin{figure*}[thb]
	\vspace{2pt}
	\setlength\tabcolsep{8pt}
	\renewcommand{\baselinestretch}{1.0}
	\footnotesize
	\centering
	\begin{subtable}[b]{1.0\linewidth}
		\centering
		\includegraphics[width=1.0\textwidth]{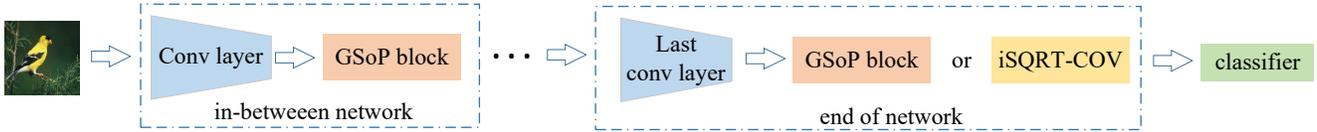}
		\caption{Overview of GSoP-Net. The proposed global second-order pooling (GSoP) block can be conveniently  inserted after any convolutional layer in-between network. We propose to use, at the network end,  GSoP block followed by common global average pooling producing compact image representations (GSoP-Net1), or  matrix power normalized covariance~\cite{Li_2017_ICCV} outputting  covariance matrices as image representations  (GSoP-Net2). \vspace{4pt}}
		\label{subfigure:overview}
	\end{subtable}
	\begin{subtable}[b]{0.95\linewidth}
		\centering
		\includegraphics[width=0.8\textwidth]{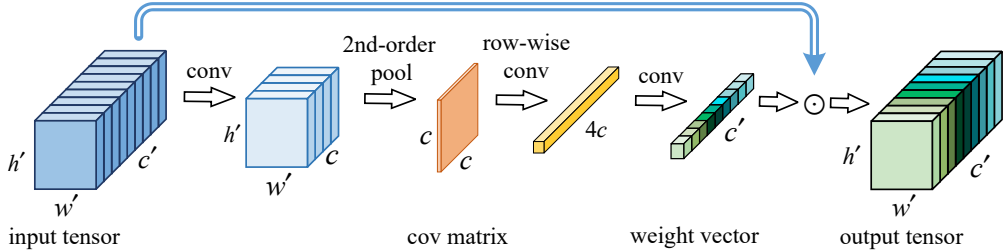}
		\caption{GSoP block. Given an input tensor, after dimension reduction, the GSoP block starts  with covariance matrix computation, followed by two consecutive operations of a linear convolution and non-linear activation, producing the output tensor which is scaling (multiplication) of the input one  along the channel dimension. }
		\label{subfigure:GSoP-block}
	\end{subtable}
	\caption{Our global second-order pooling network (GSoP-Net). Figure~\ref{subfigure:overview} gives an overview of GSoP-Net and the proposed GSoP block is presented in Figure~\ref{subfigure:GSoP-block}.  We introduce global second-order pooling into  intermediate  layers of deep ConvNets, which goes beyond the existing works where GSoP can only be used at network end. By modeling  higher-order statistics of holistic images at earlier stages, our network can enhance capability of  non-linear representation learning of deep networks.}
		\label{figure:GSoP-Net}
\end{figure*}

\paragraph{Qaudratic Transformation Network.}

The conventional network depends heavily  on linear convolution operations. Several researchers take a step further to explore higher-order transformation  for enhancing non-linear modeling  capability of deep networks. The second-order Response Transform (SORT)~\cite{Wang_2017_ICCV} develops a two-branch network module to combine responses of two  convolutional blocks and  multiplication of the responses. They perform element-wise square root for normalizing the second-order term. 
In~\cite{LiYanghao_2017_ICCV}, a factorized bilinear network (FBN) is proposed to model the pairwise feature interaction. By constraining  the rank of quadratic transformation  matrix, FBN can introduce bilinear  pooling into intermediate  layers.  Zoumpourlis et al.~\cite{Zoumpourlis_2017_ICCV} introduce Volterra kernel-based convolutions, which can model first-, second- or higher-order interactions of data, serving as approximations of non-linear functionals. All the  works above are concerned with non-linear filters, applied only to local neighborhood, just  like linear convolution. In contrast, our GSoP networks collect  the second-order statistics of the holistic image for enhancing  non-linear capability  of deep networks.

\section{Global Second-order Pooling Network}\label{section:proposed}

We illustrate the proposed GSoP-Net in Figure  ~\ref{subfigure:overview}.  Note that  the  second-order pooling block we designed  can be conveniently  inserted after any convolutional layer. By introducing this block in intermediate  layers, we can model high-order statistics of the holistic image at early stages, having ability to enhance  non-linear modeling capability of deep ConvNets.

In practice, we build two network architectures. With GSoP blocks in-between network and at the end of network, we can use GSoP block as well which is followed by the common global average pooling, producing the mean vector as compact image representation, which  we call  GSoP-Net1. Alternatively, at the end of network, we can adopt matrix power normalized covariance matrices as image representations~\cite{Li_2017_ICCV}, called GSoP-Net2, which is more discriminative yet is high-dimensional.

\subsection{Global Second-order Pooling  Block}\label{subsection:channel-block}

Figure~\ref{subfigure:GSoP-block} shows the diagram of the key module of our network, i.e., GSoP block. Similar to~\cite{Hu_2018_CVPR_SE},  the block  consists of two modules, i.e., squeeze module and excitation module. The squeeze module aims to model the  second-order statistics along the channel dimension of the input tensor. We are given a 3D  tensor of $h'\times w' \times c'$ as an input, where $h'$ and $w'$ are spatial height and width  and $c'$ is the number of channels. First, we use  $1\times 1$ convolution reducing the number of channels from $c'$ to $c$ ($c<c'$) to decrease the computational cost of the following operations. For the   $h'\times w' \times c$ tensor of reduced dimensionality, we compute pairwise channel correlations, obtaining one $c\times c$ covariance matrix. The resulting covariance matrix has clear physical meaning, i.e., its  $i^\mathrm{th}$ row indicates statistical dependency of channel $i$  with all channels. As the quadratic operations involved change  the order of data, we perform row-wise normalization for the covariance matrix, respecting the inherent structural information. In contrast, the SE-Net  uses global first-order pooling,  which can only summarize the mean  of individual channels, having limited statistical  modeling capability. 

In the excitation module, prior to channel scaling, we perform two consecutive operations of convolution plus non-linear activation for covariance matrix embedding. To maintain the structural information, the covariance matrix is subject to row-wise convolution, which is followed by a Leaky Rectified Linear Unit (LReLU). Then we perform the second convolution and this time we use the sigmoid function as a nonlinear activation, outputting  a $c\times 1$ weight vector. We finally perform dot product between the weight vector and  channels. Individual channels are thus  emphasized or suppressed in a soft manner in terms of the weights.

\subsection{Extension to Spatial Position}\label{subsection:position-block}

In previous section, we describe global second-order pooling along channel dimension, which we call \textit{channel-wise GSoP}. We can extend it to spatial position, called \textit{position-wise GSoP}, capturing pairwise feature correlations of the holistic tensor for position-wise feature scaling. 
The design philosophy of the position-wise GSoP Block is very similar to that of the channel-wise one. We also use 1$\times $1 convolution for reducing the number of channels. Furthermore, as we are to compute  pairwise  correlations of features at all spatial positions, we adopt downsampling, decreasing the spatial size to fixed $h\times w$. So we obtain a position-wise covariance matrix of  $hw\times hw$.  Row $i$ of the covariance matrix, where $i=1, \ldots, hw$ enumerates all spatial positions, indicates statistical correlation of the $i^{\mathrm{th}}$ feature with all features. The position-wise covariance matrix is also fed to two consecutive operations, i.e., row-wise  convolution$+$LReLU and convolution$+$sigmoid. After appropriate reshaping, we can obtain a  $h\times w$ weight matrix which encodes nonlinear pair-wise dependency among features  at all positions. At last, the weight matrix is upsampled to $h'\times w' \times c'$ and then multiplied  element-wise with spatial features.

\begin{figure}[tb]
	\centering
		\begin{minipage}[b]{0.7\linewidth}
			\centering
			\includegraphics[width=1.0\textwidth]{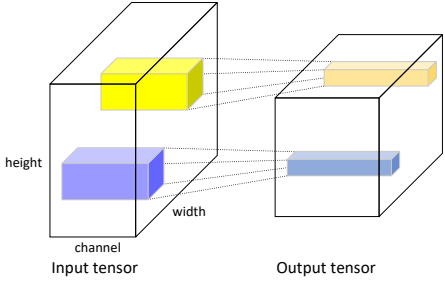}
	\end{minipage}
\caption{Classical convolutional operations fail to capture holistic dependency of 3D tensor due to limited receptive field size. For example, the data in  small blue tensor cannot interact with that of  yellow tensor at distant position due to limited receptive filed size. Our GSoP-Net addresses this by  modeling  pairwise correlations  of the holistic tensor.}
		\label{figure:insight}
\end{figure}

\subsection{Mechanism  of GSoP Block}\label{subsection-insight}

In classical deep ConvNets, restricted by limited receptive field size, the convolution operations can only process a local neighborhood of 3D tensor. The data at distant position cannot interact, e.g., the  small blue tensor and the small yellow one  as shown in Figure~\ref{figure:insight}. The long-range dependencies can only be captured by larger receptive fields produced by deep stacking of convolutional operations. This leads to  several downsides such as optimization difficulty and modeling difficulty of multi-hop dependency~\cite{Wang_2018_CVPR_nonlocal}.  

By computing all pairwise feature  correlations (or inner product), the non-local operation can capture  dependency of features  at distant positions. As a result, the non-local operation can excite significant  features, which is consistent with self-attention  machinery~\cite{NIPS2017_7181}. Our \textit{position-wise GSoP} multiplies each feature  with one weight, which encodes nonlinear correlations of this feature with features at all positions. As such, our position-wise GSoP can also model long-range dependency of  features, functioning as a kind of spatial self-attention. Beyond that, our \textit{channel-wise GSoP} can capture long-range dependency along  channel dimension, steering self-attention to significant  channels. Note that SE-Net can capture long-range channel dependency  as well, which, however, can model only the  first-order statistical dependency, having limited representation capability.

\subsection{Block Implementation}\label{subsection:block-implementation}

Our blocks can be conveniently  inserted into ResNet architecture. The ResNet contains 4 residual stages, i.e., conv2\_x, $\ldots$, conv5\_x, each containing stacks of bottleneck blocks. The exception is the first stage (i.e., conv1) which only contains  one single convolutional layer, without  bottleneck structure. To simplify block design and to tradeoff between computational complexity and classification accuracy,  we adopt fixed size covariance matrices for  all residual stages. In practice, we reduce the number of channel to 128 for both channel-wise and position-wise GSoP; in addition, we set the size of spatial covariance matrix to 64 (i.e., $h$=$w$=8). We note that the value of covariance matrix size is evaluated in Section~\ref{subsection:ablation}.

\begin{table}[t]
	\centering
	\setlength{\tabcolsep}{2pt}
	\footnotesize
	\renewcommand\arraystretch{1.3}
	\begin{tabular}{l|l|l|l|l}
		\hline 
		& \multicolumn{2}{|c|}{\parbox{1.22in}{\centering \vspace{1mm}channel-wise GSoP \vspace{1mm}}} & \multicolumn{2}{c}{\parbox{1.22in}{\centering \vspace{1mm}position-wise GSoP\vspace{1mm}}}\\
		\hline
		layers & 3D filter & output tensor  & 3D filter & output tensor\\
		\hline
		\parbox{0.5in}{\vspace{1mm} conv + BN\\ + ReLU\vspace{1mm}} & \parbox{0.5in}{1$\times$1$\times$1024\\G=1} & 14$\times$14$\times$128 & \parbox{0.5in}{1$\times$1$\times$1024\\G=1} & 14$\times$14$\times$128 \\
		\hline
		\parbox{0.7in}{\vspace{1mm}down sampling\vspace{1mm}} & \parbox{0.5in}{\centering --} & \parbox{0.5in}{\centering --}  & \parbox{0.5in}{\centering --} & 8$\times$8$\times$128\\
		\hline
		COV pool+BN & \parbox{0.5in}{\centering --} & \parbox{0.5in}{\vspace{1mm}{\color{graycolor}{128$\times$128$\rightarrow$}} \\ 1$\times$128$\times$128\vspace{1mm}} & \parbox{0.5in}{\centering --} &  \parbox{0.5in}{\vspace{1mm}{\color{graycolor}{64$\times$64$\rightarrow$}} \\ 1$\times$64$\times$64\vspace{1mm}}\\
		\hline
		\parbox{0.6in}{conv + BN + \\ LReLU (0.1)} & \parbox{0.55in}{\vspace{1mm}1$\times$128$\times $1\\G=128\vspace{1mm}} & 1$\times$1$\times$512 & \parbox{0.55in}{\vspace{1mm}1$\times $64$\times$1 \\ G=64\vspace{1mm}} & 1$\times$1$\times$256 \\ 
		\hline
		conv + sigmoid & \parbox{0.5in}{\vspace{1mm}1$\times$1$\times$512 \\G=1\vspace{1mm}} & 1$\times$1$\times$1024 & \parbox{0.5in}{1$\times$1$\times$256\\G=1} & \parbox{0.5in}{{\color{graycolor}{1$\times$1$\times$64$\rightarrow$}}\\8$\times$8$\times$1} \\
		\hline
		\parbox{0.6in}{\vspace{1mm}up sampling\vspace{1mm}} & \parbox{0.5in}{\centering --} & \parbox{0.5in}{\centering --}  & \parbox{0.5in}{\centering --} & 14$\times$14$\times$1\\
		\hline
		\parbox{0.6in}{\vspace{1mm} dot product \vspace{1mm}} & \parbox{0.5in}{\centering --} & 14$\times$14$\times$1024 & \parbox{0.5in}{\centering --} & 14$\times$14$\times$1024 \\
		\hline
		\hline
		parameters (M) &  \multicolumn{2}{|c|}{0.72} & \multicolumn{2}{c}{0.16} \\
		\hline
		MFLOPs & \multicolumn{2}{|c|}{28.1} & \multicolumn{2}{c}{26.2} \\
		\hline
	\end{tabular}%
	\vspace{2pt}
	\caption{GSoP blocks for conv4\_x.  `G' indicates \#grouped convolutions~\cite{Krizhevsky2012ImageNet}, in which  G=1 indicates common convolution (no group); gray text indicates reshape operation. Shortcut connections are added after GSoP blocks.}
	\label{tab:implemention}%
\end{table}

After the 1$\times$1 convolution for dimensionality reduction of channels, we perform downsampling   for position-wise GSoP to obtain feature maps of fixed size (i.e., $8\times 8$).
By reshaped  to a 3D tensor with first dimension being singleton, the $d$$\times$$d$ covariance matrix  can be seen as 1$\times$$d$ feature map with $d$ channels, and so row-wise BN and row-wise grouped convolutions~\cite{Krizhevsky2012ImageNet} can be easily  accomplished.  The channel number after the row convolution is raised to $4d$ and $4hw$ for channel-wise pooling and position-wise pooling, respectively.  The size of weight vector for channel-wise pooling or weight matrix for position-wise pooling, should match the input tensor size. We mention that after the proposed blocks, we  also use a shortcut connection, adding the input tensor to the scaled, output one. In Table~\ref{tab:implemention}, we present  implementation of  GSoP block  for conv4\_x.

\section{Experiments}\label{section:experiments}

In this section, we first conduct ablation analysis of the proposed GSoP-Nets. We then make comparison with the competing methods as well as state-of-the-arts on ImageNet. We finally  evaluate generalization capability  of our network to small-scale classification. All of our program are implemented under the PyTorch framework, and runs on four  workstations each of which is equipped with  2 GTX 1080Ti GPUs and an Intel i7-4790K@4GHz CPU.

\vspace{-10pt}\noindent\paragraph{Datasets} Our experiments are mainly conducted on  ImageNet-1K~\cite{imagenet_cvpr09} benchmark. The ImageNet-1K contains 1.28M training images  and 50K validation images from 1,000 classes.  In Section~\ref{subsection:ablation}, for the purpose of faster ablation study, we build a small subset of ImageNet-1K by randomly selecting 250 classes, including 320K$/$12.5K images for  training$/$validation, which we call ImageNet-$\frac{1}{4}$K. For comparison with state-of-the-art networks, we adopt standard ImageNet-1K in Section~\ref{subsection:imagenet-1k}. To evaluate the generalization  capability of our network, we also make experiments on  CIFAR-100 benchmark~\cite{CIFAR}, which   contains 60K color images of 32x32 pixels from 100 categories, with 50K images for training and 10K images for testing. 

\vspace{-10pt}\noindent\paragraph{Experimental Setting}

During training from scratch  with ResNet architecture on ImageNet, we follow~\cite{He_2016_CVPR} for data augmentation  involving scale, color and flip jittering. The weights are initialized as in~\cite{He_2015_ICCV}. We randomly crop  $224\times 224$ images from the rescaled images with per-channel mean subtraction. The networks are optimized using stochastic gradient descent (SGD) with a momentum of 0.9 and a mini-batch of 160. The initial learning rate is set to 0.1, divided by 10 every 30 epochs until 100 epochs, unless specified otherwise. During testing stage, we evaluate the error on the single   $224\times 224$   center crop from an image whose shorter size is 256. 

For training  from scratch on CIFAR-100, following g \cite{He2016_ECCV,Hu_2018_CVPR_SE}, we use  standard data augmentation for training, including horizontal flip and random translation. The networks are trained within 110 epochs with the initial learning rate of 0.25,  which is reduced to 0.025 and 0.0025 at the $80^{\mathrm{th}}$ and $95^{\mathrm{th}}$ epoch, respectively. The mini-batch size is  128 and weight decay is 1e-4.

\vspace{-2pt}\subsection{Ablation analysis on GSoP-Nets}\label{subsection:ablation}

We  develop a lightweight residual network of 26 layers (i.e., ResNet-26) as our baseline architecture, where every residual stage contains two bottlenecks. Following~\cite{Li_2017_ICCV}, we do not perform downsampling at this stage as small number of features is harmful for robust covariance estimation.  For conv2\_x$\sim$conv4\_x, we insert per-stage GSoP block   after the last bottleneck structure of each residual stage. For \textit{GSoP-Net1} we insert one GSoP block followed by common global average pooling, outputting an 2K-dimensional image representation  fully connected to softmax  layer, while for  \textit{GSoP-Net2} we use matrix power normalized covariance pooling, producing 32K-dimensional image representation. Table~\ref{tab:architectures} presents the architecture of our GSoP-Nets.

\begin{table}[t]
	\centering
	\footnotesize
	\setlength{\tabcolsep}{2pt}
	\renewcommand\arraystretch{1.1}
	\begin{tabular}{c|c|c}
		\hline
		&  output & layer\\
		\hline
		conv1 & 112$\times$112 & conv, 7$\times$7, 64, Stride$=$2  \\
		\hline
		pool1 & \multirow{4}{*}{56$\times$56} & max pool, 3$\times$3, Stride$=$2 \\
		\cline{1-1}\cline{3-3}
		conv2\_x & & \parbox{1.1in}{\vspace{1mm}\centering $\left[
			\begin{matrix}
			\mathrm{conv},1\times1,64 \\
			\mathrm{conv},1\times1,64 \\
			\mathrm{conv},1\times1,256
			\end{matrix}
			\right]\times 2$ \\ 
			GSoP Block\vspace{1mm}}  \\
		\hline
		conv3\_x &28$\times$28 &\parbox{1.1in}{\vspace{1mm}\centering  $\left[
			\begin{matrix}
			\mathrm{conv},1\times 1,128 \\
			\mathrm{conv},1\times 1,128 \\
			\mathrm{conv},1\times 1,512
			\end{matrix}
			\right]\times 2$\\ 
			GSoP Block\vspace{1mm}}     \\
		\hline
		conv4\_x & 14$\times$14&\parbox{1.15in}{\vspace{1mm}\centering $\left[
			\begin{matrix}
			\mathrm{conv},1\times 1,256 \\
			\mathrm{conv},1\times 1,256 \\
			\mathrm{conv},1\times 1,1024
			\end{matrix}
			\right]\times2$\\ 
			GSoP Block\vspace{1mm}}    \\
		\hline
		conv5\_x &14$\times$14 &\parbox{1.15in}{\vspace{1mm}\centering $\left[
			\begin{matrix}
			\mathrm{conv},1\times 1,512 \\
			\mathrm{conv},1\times 1,512 \\
			\mathrm{conv},1\times 1,2048
			\end{matrix}
			\right]\times 2$\\ 
		}     \\
		\hline
		& 1$\times 1$ & {\parbox{1.8in}{\centering \vspace{4pt}GSoP block+GAvP, 2K  \\ \quad \quad or \\ iSQRT-COV~\cite{Li_2018_CVPR}, 32K\vspace{4pt}}}  \\
		\hline
		&  1$\times 1$ & FC + softmax\\
		\hline
	\end{tabular}%
	\caption{GSoP-Net with ResNet-26 architecture.}\label{tab:architectures}%
\end{table}

\begin{table}[htb!]
	\centering
	\setlength{\tabcolsep}{2pt}
	\footnotesize
	\renewcommand\arraystretch{1.3}
	\begin{tabular}{c}
		\begin{minipage}{1\linewidth}
			\begin{subtable}{1\linewidth}
				\centering
				\footnotesize
				\setlength{\tabcolsep}{6pt}
				\begin{tabular}{c|c|c|c}
					\hline
					\multicolumn{2}{c|}{\multirow{2}{*}{}} & \multicolumn{2}{c}{top-1 err$/$top-5 err}\\
					\cline{3-4}
					\multicolumn{2}{c|}{}  &GSoP-Net1 &GSoP-Net2\\ 
					\hline
					\multirow{3}{*}{\parbox{0.6in}{\centering channel-wise\\cov size $c$}}& 64$\times$64 & 18.00$/$4.99 & 16.84$/$4.58 \\						
					& 128$\times$128 & $\textbf{17.42}$/$\textbf{4.53}$ & \textbf{16.68}$/$\textbf{4.36} \\
					& 256$\times$256 & 17.61$/$4.64  & 16.67$/$4.18 \\
					\hline
					\multirow{3}{*}{\parbox{0.6in}{\centering position-wise\\cov size $hw$}}& 36$\times$36 & 19.21$/$5.46 & 17.34$/$4.80 \\						
					& 64$\times$64 & \textbf{18.37}$/$\textbf{5.05} & \textbf{17.18}$/$\textbf{4.80} \\
					& 144$\times$144 & 18.41$/$5.08 & 17.51$/$4.63 \\
					\hline
					\multicolumn{2}{c|}{vanilla network}  & \multicolumn{2}{c}{19.18$/$5.62}\\	
					\hline		
				\end{tabular}%
				\setlength{\abovecaptionskip}{1.5pt}
				\setlength{\belowcaptionskip}{4pt}
				\caption{Impact of covariance matrix  size.}\label{subtab:cov-size}
				
			\end{subtable}
		\end{minipage} \\
		
		\begin{minipage}{1\linewidth}
			\begin{subtable}{1\linewidth}
				\centering
				\footnotesize
				\setlength{\tabcolsep}{5pt}
				\begin{tabular}{p{0.6cm}<{\centering}|p{1.5cm}<{\centering}|p{1.7cm}<{\centering}|p{1.7cm}<{\centering}}
					\hline
					\multicolumn{2}{c|}{\multirow{2}{*}{}}& \multicolumn{2}{c}{top-1 err$/$top-5 err}\\
					\cline{3-4}
					\multicolumn{2}{c|}{}&GSoP-Net1 &GSoP-Net2\\ 
					\hline
					\multicolumn{2}{c|}{channel-wise pool}& \textbf{17.42}$/$\textbf{4.53} & \textbf{16.68}$/$\textbf{4.36} \\
					\multicolumn{2}{c|}{position-wise pool}& 18.37$/$5.05 &17.18$/$4.80 \\
					\hline												
					\multirow{3}{*}{fusion}& average & 17.90$/$4.73 & 16.77$/$4.36  \\						
					& maximum & \textbf{17.48}$/$\textbf{4.52} & 16.80$/$4.39\\
					& concatenation &17.58$/$4.61 & \textbf{16.49}$/$\textbf{4.35} \\						
					\hline
				\end{tabular}%
				\setlength{\abovecaptionskip}{1.5pt}
				\setlength{\belowcaptionskip}{4pt}
				\caption{Comparison of fusion schemes. }\label{subtab:fusion}
			\end{subtable}
		\end{minipage} \\
		
		\begin{minipage}{1\linewidth}
			\begin{subtable}{1\linewidth}
				\centering
				\footnotesize
				\setlength{\tabcolsep}{13pt}
				\begin{tabular}{l|c|c}
					\hline 
					$[\mathrm{S}2, \mathrm{S}3, \mathrm{S}4, \mathrm{S}5]$ & top-1 err & top-5 err  \\
					\hline
					$[-,\;-,\;-,\;-\;]$ & 19.18 & 5.62  \\
					$[\mathrm{C},\;-,\;-,\;-\;]$ & 18.45 & 5.22   \\
					$[-,\;\mathrm{C},\;-,\;-\;]$ & 18.72 & 5.33   \\
					$[-,\;-,\;\mathrm{C},\;-\;]$ & 18.85 & 5.24  \\
					$[-,\;-,\;-,\;\mathrm{C}\;]$ & 18.33 & 5.12   \\
					\hline
					$[\mathrm{C},\;\mathrm{C},\;\mathrm{C},\;\mathrm{C}\;]$ & 17.42 & 4.53   \\
					\hline
					$[-,\;-,\;-,\;i\sqrt{\;}]$ & 17.43 & 4.71   \\
					$[\mathrm{C},\;\mathrm{C},\;\mathrm{C},\;i\sqrt{\;}]$ & \textbf{16.68} & \textbf{4.36}   \\
					\hline
				\end{tabular}%
				\setlength{\abovecaptionskip}{1.5pt}
				\setlength{\belowcaptionskip}{4pt}
				\caption{Single block performance. }\label{subtab:stage_comparison}
			\end{subtable}%
		\end{minipage}\\
	\end{tabular}
	\caption{Ablation results of our GSoP-Nets with ResNet-26 architecture on ImageNet-$\frac{1}{4}$K.}
	\label{tab:ablation_results}
\end{table}

\vspace{-10pt}\paragraph{Impact of Covariance Size.}

The covariance matrices, produced by the second-order pooling blocks, encode the statistical correlation of the holistic tensors, playing a central role in our networks. So we first evaluate impact of covariance matrix size on the proposed  networks.  Table~\ref{subtab:cov-size} summarizes the  results, in which the top and middle panel shows the impacts using channel-wise (cov size: $c\times c$)  and position-wise pooling (cov size: $hw\times hw$), respectively.  We first observe that, whatever the second-order pooling, the proposed networks improve over vanilla  ResNet-26, demonstrating that  our holistic modeling methods in earlier stages are beneficial  in enhancing the network's discriminative capability.  For channel-wise second-order pooling, relative to varying values of $c$,  GSoP-Net1 achieves the best results with $c=128$. The errors of GSoP-Net2 consistently decline as $c$ gets larger and the lowest error is obtained with $c=256$. For position-wise second-order pooling,  GSoP-Net1 with $hw=64$ produces the lowest errors.   Notably, for either channel-wise or position-wise  pooling, it is clear that GSoP-Net2 performs much better than GSoP-Net1, which suggests that image representation of covariance matrix is superior to that of mean vector by average pooling.

\paragraph{Fusion of Channel- and Position-wise  Pooling.}

The channel-wise  and position-wise second-order pooling capture statistical correlations from different dimensions of 3D tensor. They can be combined for holistic image modeling. Given an input tensor, we  independently perform second-order poolng along the channel dimension and spatial dimension, producing two output tensors. We can fuse the two output tensors by the commonly used operations of average$/$maximum  and concatenation. As concatenation operation increases tensor size, we use one convolutional layer for maintaining the original tensor size.

The  results of fusion methods are presented in Table~\ref{subtab:fusion}. For GSoP-Net1, the average scheme performs worse than the other two, while the maximum scheme is slightly better than the concatenation one. For GSoP-Net2, the concatenation scheme is a little superior to the other two schemes. However, compared to separate channel-wise  pooling,  with any fusion scheme, combination of position-wise pooling  brings little improvement. These results suggest that the two kinds of  second-order pooling methods are not complementary, though the two proposed networks individually have  obvious improvement over the vanilla network. 

\vspace{-8pt}\paragraph{Performance of Single Second-order Block.}

In this part, we conduct experiments to analyze the performance of single channel-wise block separately  added to different residual stage. We neglect analysis on single position-wise block as it is not promising. Table~\ref{subtab:stage_comparison} presents the results, where S2 denotes residual stage 2, and so on; $-$,  C and $i\sqrt{\;}$ denote no second-order block, one channel-wise block and iSQRT-COV meta layer~\cite{Li_2017_ICCV} inserted at the corresponding residual stage, respectively. It can be seen that  insertion of single block into any residual stage brings comparable improvement over the vanilla  network. This indicates the channel-wise second-order pooling at different stage makes similar contribution to the overall performance of channel-wise GSoP-Net1. The iSQRT-COV, which inserts a matrix normalized covariance matrix at residual stage 4 as the final image representation, is a strong baseline, achieving comparable result with GSoP-Net1. The GSoP-Net2, which inserts  global second-order pooling at intermediate stages, outperform iSQRT-COV by a non-trivial margin. This suggests the benefits of introducing second-order statistics in earlier layers of networks.

\subsection{Results  on ImageNet-1K}\label{subsection:imagenet-1k}

\begin{figure}[t]
	\setlength\tabcolsep{4pt}
	\footnotesize
	\centering
	\begin{subfigure}[t]{0.7\linewidth}
		\centering
		\includegraphics[width=1\textwidth]{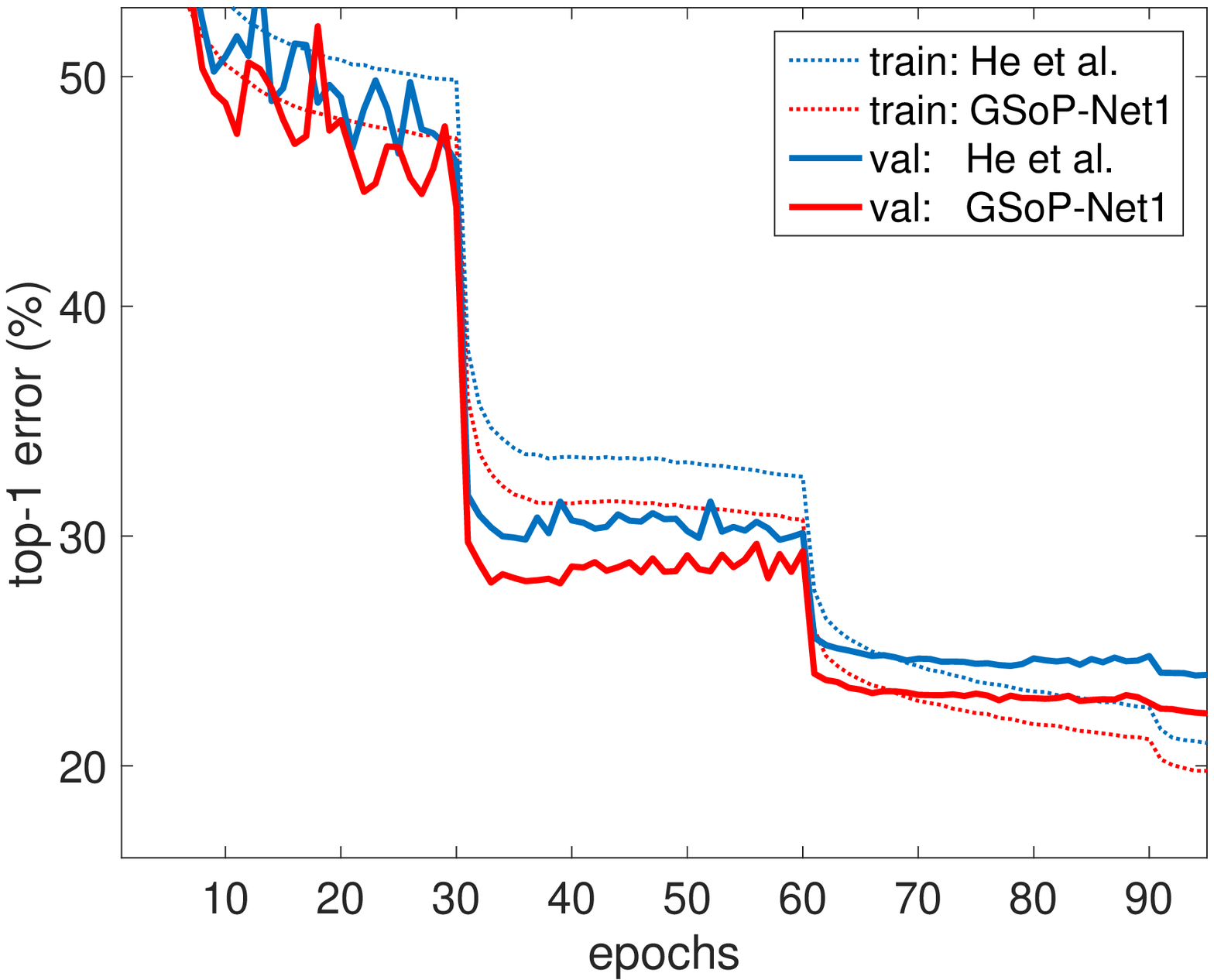}
	\end{subfigure}
	\\
	\begin{subfigure}[t]{0.7\linewidth}
		\centering
		\includegraphics[width=1\textwidth]{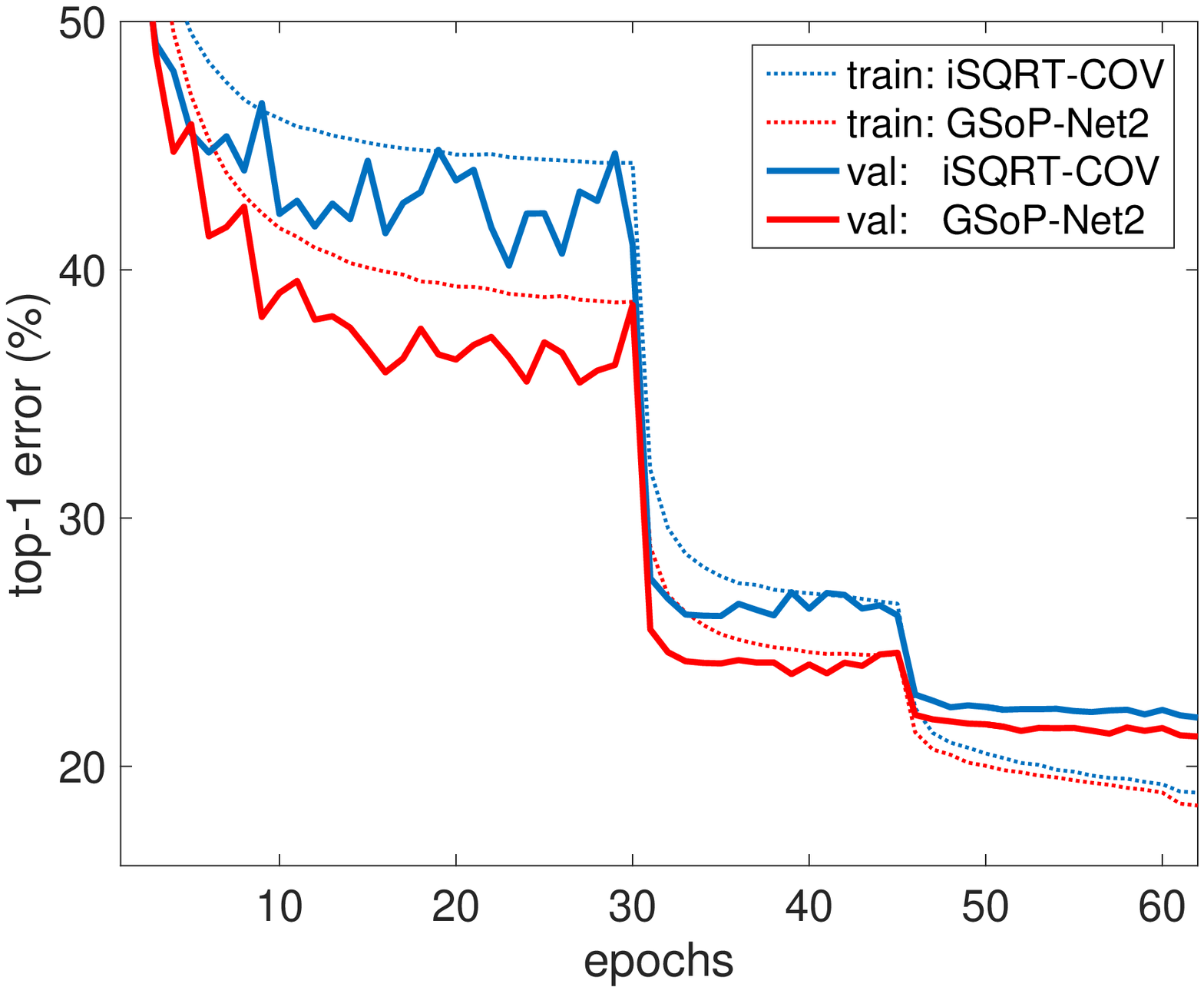}
	\end{subfigure}
	\caption{Convergence curves of our GSoP-Nets under ResNet-50 architecture. Top: GSoP-Net1 vs vanilla network; bottom: GSoP-Net2 vs iSQRT-COV.}
	\label{figure:curve}
\end{figure}

In this subsection, we further evaluate our proposed GSoP-Nets on standard ImageNet-1K under ResNet-50 architecture.  We insert a GSoP block  for residual stage 2, 3 and 4, respectively. For GSoP-Net1, we insert one GSoP block for residual  stage 5, followed by the commonly used global average pooling; for GSoP-Net2, instead of the GSoP block, the meta-layer of iSQRT-COV~\cite{Li_2018_CVPR}  is inserted. 

\subsubsection{Convergence and Network Complexity} 

\vspace{-2pt}\paragraph{Convergence.} Figure~\ref{figure:curve} illustrates the convergence curves  of our GSoP-Net. For GSoP-Net1, though second-order statistical modeling is exploited, it is  for tensor scaling while the convolutional filters and image representation are of both linear just like the original ResNet-50. As shown in the top figure, the convergence behavior of GSoP-Net1  is similar to that of ResNet-50, but consistently has lower validation error throughout the training process. Different from iSQRT-COV, for GSoP-Net2 we introduce  second-order blocks for residual stages 1,2 and 3. From the bottom Figure, we can see that GSoP-Net2 inherits fast convergence property of iSQRT-COV, while steadily performs better.  We attribute the improvement of our networks over their counterparts  to the holistic modeling of second-order statistics introduced in earlier stages. 

\vspace{-8pt}\paragraph{Network Complexity.}  Table~\ref{tab:ImageNet} shows  comparison of parameter and computation.  The number of parameters of GSoP-Net1 is comparable to that of the vanilla ResNet-50, while GSoP-Net2 has nearly doubled  the number of parameters. The increased parameters in GSoP-Net2 are mainly due to FC layer, in which dimensionality of image representation is 32K, accounting for most increase of the total parameters, just like MPN-COV~\cite{Li_2017_ICCV} and iSQRT-COV~\cite{Li_2018_CVPR}. We argue that advances on model compression, e.g., \cite{denton2014exploiting,Rastegari2016XNOR,han2015learning},  has potential to significantly  reduce the number of  parameters, particularly in FC layer,  while maintaining the performance; in practice, we can exploit such techniques  for reducing parameters. Analogous  to~\cite{Li_2017_ICCV,Li_2018_CVPR}, the GFLOPs of our  networks are 1.58x of the number of vanilla  ResNet. The computations increased are attributed to removal of downsampling  in the last residual stage, so that feature map size doubles. This operation is  helpful for robust covariance estimation by alleviating the problem of small sample and high dimensionality~\cite{Li_2017_ICCV}. This somewhat slowdowns the training, however, while making little difference for  inference. With a single GTX 1080Ti GPU with CUDA 9.0 and CuDNN7.1, the inference time (ms) per image are 2.52 vs 2.68$/$2.84 (vanilla ResNet-50 vs GSoP-Net1$/$GSoP-Net2). 

\begin{table}[tb!]
	\centering
	\setlength{\tabcolsep}{3pt}
	\footnotesize
	\renewcommand\arraystretch{1.3}
	\begin{tabular}{l|c|c|c|c}
		\hline
		&	description &  top-1$\;$  & top-5$\;$& params/GFLOPs \\ 
		\hline
		He et al.~\cite{He_2016_CVPR} &\parbox{0.8in}{\vspace{1mm}Baseline network\vspace{1mm}} &23.85&7.13 & 25.5M/3.86\\
		\hline
		FBN~\cite{LiYanghao_2017_ICCV} & \multirow{2}{*}{\parbox{0.8in}{Quadratic transformations}}& 24.0&7.1 & --\\
		SORT~\cite{Wang_2017_ICCV} & & 23.82& 6.72& --  \\
		\hline
		MPN-COV~\cite{Li_2017_ICCV} & \multirow{2}{*}{\parbox{0.8in}{GSoP at network end}}& 22.74&6.54 &2.2$\times$/1.6$\times$\\
		iSQRT-COV~\cite{Li_2018_CVPR} &  &22.14&6.22 &  2.2$\times$/1.6$\times$ \\
		\hline
		SE-Net~\cite{Hu_2018_CVPR_SE} & \multirow{3}{*}{\parbox{0.8in}{GAvP across network}}& 23.29&6.62 & 1.1$\times$/1.0$\times$\\
		GENet~\cite{hu2018genet} & & 21.88&5.80 & 1.3$\times$/1.0$\times$\\
		CBAM~\cite{Woo_2018_ECCV} &	& 22.66&6.31& 1.1$\times$/1.0$\times$\\
		\hline
		GSoP-Net1 (ours) &	\multirow{2}{*}{\parbox{0.8in}{GSoP across network}}& 22.32&6.02 & 1.1$\times$/1.6$\times$\\
		GSoP-Net2 (ours) & 	& \textbf{21.19}&\textbf{5.64} & 2.3$\times$/1.7$\times$ \\
		\hline
		\hline
		ResNeXt~\cite{Xie_Ross_2017_CVPR} &\multirow{3}{*}{\parbox{0.8in}{Modified architectures upon ResNet}}& 22.11 &5.90 & 1.0$\times$/1.0$\times$\\ 		DropBlock~\cite{ghiasi2018dropblock} & &21.87&5.98 & 1.0$\times$/1.0$\times$\\ 
		DRN-A-50~\cite{yu2017dilated} & & 22.94& 6.57 & 1.0$\times$/4.9$\times$\\
		\hline
		
	\end{tabular}%
	\caption{Comparison (\%) of different methods with ResNet-50 architecture on ImageNet-1K.}
	\label{tab:ImageNet}%
\end{table}

\subsubsection{Comparison with Competing Networks.}

Table~\ref{tab:ImageNet} compares  classification errors between our GSoP-Nets and the competing networks  on ImageNet-1K. 

\vspace{-6pt}\paragraph{Comparison with FBN and SORT} The two works~\cite{LiYanghao_2017_ICCV,Wang_2017_ICCV} are among the first which introduce quadratic transformation, instead of just linear convolutions,   throughout a network. However, compared to the vanilla network, their performance gains are not significant. In contrast, our networks are much better, achieving over 2.8\% and 2.6\% higher accuracies than FBN and SORT. This comparison demonstrates that,  by making favorable use of higher-order information, we can greatly improve the network performance. 

\vspace{-6pt}\paragraph{Comparison with Global Cov Pool at Network End.} Here we compare our GSoP-Net2 with several methods where global second-order pooling is inserted only at the end of network. All of them estimate covariance matrices of the last convolutional features as image representations.  DeepO$_2$  computes matrix logarithm for covariance matrix while B-CNN performs element-wise power normalization plus $\ell_2$ normalization. As  DeepO$_2$ and B-CNN are not competitive for large-scale visual recognition~\cite{Li_2017_ICCV}, here we do not compare with them. MPN-COV uses structured normalization by matrix square root, and iSQRT-COV is a faster version of MPN-COV, in which matrix square root is based on iterative algorithm, rather than GPU unfriendly SVD. 
Our GSoP-Net2 outperforms MPN-COV by 1.55\% in top-1 error (0.90\% in top-5 error). Compared to iSQRT-COV, the GSoP-Net2 achieves 0.95\%$/$0.58\% lower  top-1$/$top-5 error rates, while resulting in negligible  overhead. We note that the iSQRT-COV is a strong baseline and our improvement is nontrivial. The comparison between our GSoP-Net2 and MPN-COV$/$iSQRT-COV indicates that introducing higher-order statistics in earlier stages can enhance representational  learning capability of deep ConvNets.

\vspace{-6pt}\paragraph{Comparison with Global Avg Pool across Network.}   From Table~\ref{tab:ImageNet}, we can see that our GSoP-Net1 performs $1.0\%/0.6\%$ better than SE-Net in  top-1$/$top-5 errors. As an extension of SE-Net, CBAM combines global average and max pooling along both channel dimensional and spatial dimension. Nevertheless, the error rates of GSoP-Net1 are lower than CBAM. Building upon SE-Net, GENet~\cite{hu2018genet} proposes gather and excitation operations for exploiting context information. Our GSoP-Net2 outperforms GENet by a non-trival margin.  These  comparisons between our networks and SE-Net and its variants show that higher-order  modeling is able to capture richer statistics than the first-order modeling, leading to more discriminative representation. Notably,  we do not insert  GSoP block after each bottleneck structure; instead, we only insert the GSoP block per residual stage. As a result, we only add no more than 4 GSoP blocks, and more GSoP blocks may further improve the performance of our network.  

\vspace{-10pt}\paragraph{Comparison with State-of-the-arts.} 
Finally, we compare with several state-of-the-art networks which modify upon ResNet-50 architecture. Compared to ResNet,  ResNeXt~\cite{Xiao_2018_CVPR} considerably increases network width, which, however, keeps parameters and computation almost unchanged through extensive use of grouped convolutions~\cite{Krizhevsky2012ImageNet}. DRN-A-50~\cite{yu2017dilated}  removes downsampling in residual stage 3 and 4, and meanwhile uses dilated convolution to maintain the receptive size.  DropBlock~\cite{ghiasi2018dropblock}  extends dropout technique to  convolution; by drop  blocks of feature map randomly, it  maintains the context integrity during training. As shown in Table~\ref{tab:ImageNet},  these modified networks performs much better than ResNet-50. Nevertheless, our GSoP-Net2 outperforms all of them by a non-trivial  margin. It is noteworthy to mention that, if built upon the modified networks above,  the performance of our network may improve further.

\subsection{Results on CIFAR-100}

This section conducts experiments on CIFAR-100~\cite{CIFAR} to evaluate the generalization capability of the proposed GSoP-Net.  The backbone network is pre-activation ResNet-164~\cite{He2016_ECCV}, containing 3 residual stages each with 18 bottlenecks; the final image represenation is 256-D. In GSoP-Net1, we insert 18 GSoP blocks into the backbone network uniformly, and in GSoP-Net2 the last GSoP block is replaced by a meta-layer of iSQRT-COV.  Downsampling is not performed in the last residual stage. The final dimension of image representation in GSoP-Net2 is 8K and  a dropout layer (dropout rate=0.5) is used for FC layer. The covariance size is  $64\times 64$ in both GSoP-Net1 and GSoP-Net2.

The experimental results on CIFAR-100  are presented in Table~\ref{tab:CIFAR-100}. Compared with the vanilla  network, GSoP-Net1 and GSoP-Net2 obtain gains of 3.47\% and 5.75\%, respectively, improving the performance by a large margin.  CMPE~\cite{hu2018competitive} implements channel-wise excitation operation by establishing the correlation of the channel-wise representation between two nearby bottlenecks, which can be considered as a cross-block version of SE-Net.
GSoP-Net1 performs better than SE-Net and CMPE by 0.45\% and 1.49\% respectively.   iSQRT-COV is very competitive, outperforming SE-Net by $\sim\!\!\! 1.36\%$. By introducing second-order statistics in earlier stages, our GSoP-Net2 makes further improvement ($\uparrow$ 1.37\%) over iSQRT-COV.

\begin{table}[tb!]
	\centering
	\setlength{\tabcolsep}{2pt}
	\footnotesize
	\renewcommand\arraystretch{1.2}
	\begin{tabular}{l|c|c|c}
		\hline
		model & top-1 err & params & GFLOPs \\ 
		\hline
		He et al~\cite{He2016_ECCV}      & 24.33 & 1.7M & 0.25\\
		\hline
		SE-Net~\cite {SE_pami}           & 21.31 & 1.9M & 0.29\\	
		CMPE~\cite{hu2018competitive}    & 22.35 & 2.0M & N$/$A\\	
		MPN-COV~\cite{Li_2018_CVPR}      & 19.95 & 2.5M & 0.52\\
		\hline
		GSoP-Net1 (ours) & 20.86 & 2.9M & 0.55 \\
		GSoP-Net2 (ours) & \textbf{18.58} & 3.6M & 0.58\\
		\hline		
	\end{tabular}%
	\caption{Error comparison  (\%) of our networks with the counterparts on CIFAR-100.}
	\label{tab:CIFAR-100}%
\end{table}

\section{Conclusion}

We presented a simple yet effective deep convolutional network model for  capturing holistic statistical correlations across all stages of network. By exploiting the holistic higher-order  information at earlier  stages, the proposed model can learn more discriminative   representations. As far as we know, our work is among the first which  introduce global second-order pooling  into lower layers of deep networks. Our proposed networks performs  better than  SE-Net~\cite{Hu_2018_CVPR_SE}, i.e., the first-order counterpart, while non-trivially improves state-of-the-art iSQRT-COV~\cite{Li_2018_CVPR} which plugged global covariance pooling as image representation  only at network end.  The proposed GSoP blocks are highly modular, which  can be conveniently  plugged into other deep architectures, e.g., Inception~\cite{Szegedy_2015_CVPR} and DenseNet~\cite{Huang_2017_CVPR}.

{\small
\bibliographystyle{ieee}
\bibliography{egbib}
}

\end{document}